\documentclass{article}
\usepackage{oraclegap_arxiv,times}
\usepackage{amsmath}
\usepackage{booktabs}
\usepackage{graphicx}
\usepackage{tikz}
\usetikzlibrary{arrows.meta,calc,decorations.pathreplacing,fit,positioning}
\usepackage{microtype}
\usepackage{hyperref}
\usepackage{url}
\hypersetup{
  colorlinks=true,
  linkcolor=blue,
  citecolor=blue,
  urlcolor=blue,
  pdftitle={Oracle Gap and Signal Fidelity: A Fixed-Pool Diagnostic for Test-Time Collaboration},
  pdfauthor={Jie Hu}
}

\title{Oracle Gap and Signal Fidelity: A Fixed-Pool Diagnostic for Test-Time Collaboration}
\author{Jie Hu\\
Research Institute of China Telecom\\
\texttt{hujiesse@gmail.com}}
\iclrfinalcopy

\begin{document}
\maketitle
\fancyhead{}

\begin{abstract}
Test-time collaboration, including self-consistency, best-of-$N$ selection, critic models, and verifier pipelines, is often credited with broadly improving LLM reasoning, yet its gains are uneven and sometimes negative. We ask when training-free collaboration should be expected to help. For a fixed candidate pool, we decompose a selector or verifier's net gain into measurable factors: recoverable mass, verification-signal coverage, conditional selection quality, and harm to already-correct outputs. This reframes collaboration as a candidate-selection problem rather than as an intrinsic property of a multi-agent topology. Across LiveCodeBench, MATH Level-5 hard subjects, and GPQA-Diamond, gains are bounded first by the oracle gap and then by signal fidelity, which we measure directly as candidate-level agreement between verifier verdicts and official labels. On LiveCodeBench, a public-test verifier (MCC 0.825) gains $+8.14$ percentage points (pp) over a first-sample baseline; a generated-test verifier (MCC 0.248) improves by $+2.70$pp and is not statistically distinguishable from an LLM selector, but operates at near-zero harm versus the selector's 4.69\% harm rate. On MATH, a symbolic answer-equivalence selector beats self-consistency by $+4.67$pp, while LLM selectors are negative. On GPQA-Diamond, recoverable mass is only 3.03\% and 87.54\% of candidate pools are answer-identical; a weaker model's pools shrink both further, suggesting that oracle gap is a joint property of task, model, and sampling configuration. Our framework yields a practical pre-deployment diagnostic: estimate the oracle gap, then measure coverage, signal fidelity, and harm before investing in collaboration.
\end{abstract}

\section{Introduction}

Test-time collaboration is now a common recipe for improving large language model reasoning: sampling multiple candidates and voting by self-consistency, asking a critic to inspect an answer, filtering code with tests, or routing an input to a stronger worker. These mechanisms are often grouped under one intuition: more computation, more agents, or more judging should improve reliability.

Empirically, this intuition is incomplete. A critic may fix some failures while damaging answers that were already correct. Self-consistency is meaningful for math problems with normalized final answers, but exact majority voting over code strings is mostly a fallback rule rather than a real selector. Public tests can be an extremely strong signal for code, while self-generated tests can inherit the model's own blind spots. A second model may add diversity, or it may simply choose among nearly identical candidates, right or wrong. In short, collaboration is not a property of the topology alone: its value depends on the candidate pool, the task's evaluation structure, and the fidelity of the signal used to choose among candidates.

This paper asks: \emph{when should training-free collaboration be expected to help?} We refer to the resulting fixed-pool diagnostic framework as \emph{OracleGap}. We study the selector/verifier setting: for each problem, a fixed candidate pool is generated, and a mechanism chooses one candidate without further training. This covers best-of-$N$ selection, LLM selectors, generated-test filtering, public-test filtering, and symbolic answer-equivalence selection, and deliberately separates selection from repair: a repair mechanism can create new candidates, whereas a selector can only capture the improvement already present in the pool.

\begin{figure}[t]
\centering
\resizebox{\textwidth}{!}{\begin{tikzpicture}[
  x=1cm,
  y=1cm,
  font=\footnotesize,
  box/.style={draw=black!55, rounded corners=2pt, align=center, fill=gray!5,
    inner sep=3.5pt, minimum height=8mm},
  signal/.style={box, draw=blue!65!black, fill=blue!8},
  accounting/.style={box, draw=orange!60!black, fill=orange!8},
  decision/.style={box, draw=green!45!black, fill=green!9},
  candidate/.style={draw=black!60, minimum width=4.3mm, minimum height=4.8mm,
    inner sep=0pt, font=\scriptsize},
  good/.style={candidate, fill=green!24},
  bad/.style={candidate, fill=red!20},
  arrow/.style={-{Stealth[length=1.8mm]}, line width=0.55pt, draw=black!65},
  audit/.style={-{Stealth[length=1.6mm]}, dashed, line width=0.5pt,
    draw=black!55},
  note/.style={align=center, font=\scriptsize, text=black!68},
]

\node[anchor=west, font=\footnotesize\itshape, text=black!65]
  at (-0.05,1.12) {(a) Fixed-pool diagnostic and gain decomposition};

\node[box, minimum width=1.25cm] (input) at (0.65,0) {Input $x$};

\node[bad]  (y1) at (2.05,0) {$y_1$};
\node[good] (y2) at (2.55,0) {$y_2$};
\node[bad]  (y3) at (3.05,0) {$y_3$};
\node[good] (y4) at (3.55,0) {$y_4$};
\node[bad]  (y5) at (4.05,0) {$y_5$};
\node[draw=black!50, rounded corners=2pt, fit=(y1)(y5), inner xsep=2.5mm,
  inner ysep=2.2mm] (pool) {};
\node[font=\scriptsize, above=1.5mm of pool] {Fixed pool $C(x)$, $k=5$};
\node[font=\tiny, below=0.7mm of y1, text=black!60] {reference};

\node[signal, minimum width=3.25cm] (sig) at (6.35,0)
  {Verification signal\\[-1pt]
   {\scriptsize coverage: signal defined?}\\[-2pt]
   {\scriptsize fidelity: verdict vs. official label}};

\node[box, minimum width=1.85cm] (select) at (9.25,0)
  {Select $\hat y$\\[-1pt]{\scriptsize fallback $\to y_1$}};

\node[accounting, minimum width=3.15cm] (outcome) at (12.15,0)
  {Outcome accounting\\[-1pt]
   {\tiny fixed $(+)$ / harmed $(-)$ / unchanged $(0)$}};

\draw[arrow] (input) -- (pool);
\draw[arrow] (pool) -- (sig);
\draw[arrow] (sig) -- (select);
\draw[arrow] (select) -- (outcome);

\draw[decorate, decoration={brace,mirror,amplitude=3pt}, draw=black!55]
  ($(pool.south west)+(0,-0.42)$) -- ($(pool.south east)+(0,-0.42)$)
  node[midway, below=1.5mm, note, text width=3.4cm]
  {ref. wrong $+$ correct alternative\\
   $\Rightarrow$ recoverable; \mbox{\textbf{oracle gap}}\\[-1pt]
   {\tiny official-label audit only}};

\node[align=center, font=\scriptsize, text=black] at (9.15,-1.18)
  {$\mathrm{gain}=P(R\wedge D)\,q-P(C\wedge D)\,h$\\[-1pt]
   {\tiny $R$: recoverable, $C$: reference correct, $D$: signal defined}};

\node[anchor=west, font=\footnotesize\itshape, text=black!65]
  at (-0.05,-2.52) {(b) Pre-deployment workflow};

\node[box, text width=2.75cm, minimum height=1.05cm] (step1) at (1.55,-3.48)
  {\textbf{1. Estimate}\\oracle gap};
\node[box, text width=2.75cm, minimum height=1.05cm] (step2) at (5.05,-3.48)
  {\textbf{2. Audit}\\coverage and fidelity};
\node[box, text width=2.75cm, minimum height=1.05cm] (step3) at (8.55,-3.48)
  {\textbf{3. Compare}\\capture against harm};
\node[decision, text width=2.75cm, minimum height=1.05cm] (step4) at (12.05,-3.48)
  {\textbf{4. Deploy}\\only if net-positive};

\draw[arrow] (step1) -- (step2);
\draw[arrow] (step2) -- (step3);
\draw[arrow] (step3) -- (step4);

\node[note, text width=3.05cm] at (1.55,-4.62)
  {small gap $\Rightarrow$ stop\\\emph{GPQA: 3.03 pp}};
\node[note, text width=3.05cm] at (5.05,-4.62)
  {low coverage/fidelity limits capture\\\emph{L4-gen: 35.8\% active}\\\emph{MCC 0.248}};
\node[note, text width=3.05cm] at (8.55,-4.62)
  {harm can erase recovered gains\\\emph{GPQA L1: net $-1.68$ pp}};
\node[note, text width=3.05cm] at (12.05,-4.62)
  {high-fidelity signals pay\\\emph{L4-public: $+8.14$ pp}\\\emph{zero harm}};

\end{tikzpicture}}
\caption{OracleGap overview. (a) Official labels expose the audit-only oracle gap and measure the coverage and fidelity of a deployable signal; selection then produces fixes, harms, or no changes. (b) The pre-deployment workflow stops when recoverable mass is small and deploys only signals whose captured gains exceed harm.}
\label{fig:framework}
\end{figure}
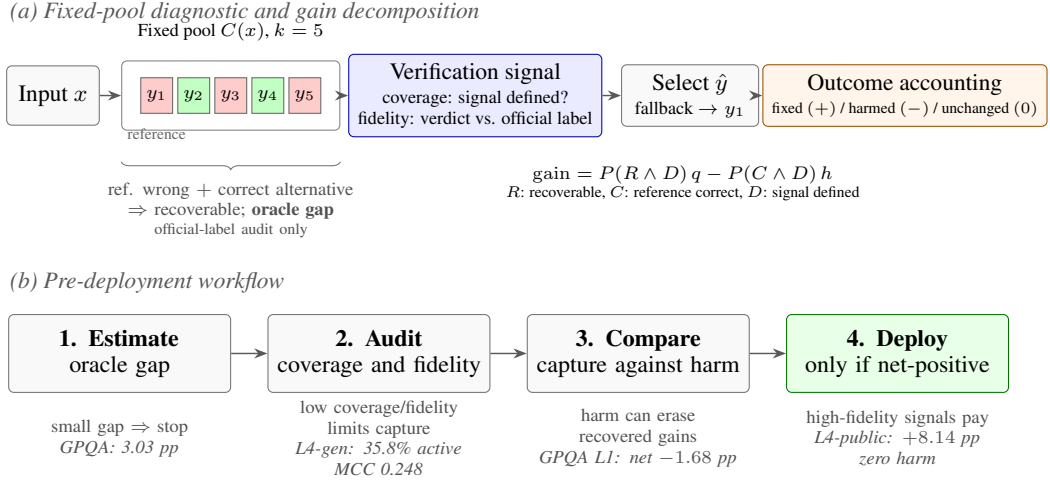

Our central claim is that selection gains are bounded first by the \emph{oracle gap} and then by the \emph{coverage, fidelity, quality, and harm} of the verification signal. We make this concrete with a fixed-pool decomposition:
\begin{equation}
\begin{split}
\mathrm{gain}={}&P(\mathrm{recoverable}\wedge\mathrm{defined})\,q\\
&-P(\mathrm{reference\ correct}\wedge\mathrm{defined})\,h,
\end{split}
\label{eq:decomposition}
\end{equation}
where \emph{recoverable} means the reference output is wrong but some candidate in the pool is correct; $q$ is conditional selection success on recoverable examples with a defined signal; and $h$ is how often the mechanism changes a reference-correct example into an incorrect one among examples with a defined signal. We also measure verifier \emph{fidelity} directly as candidate-level agreement between verifier verdicts and official labels, linking the decomposition to imperfect-verifier theory.

We evaluate on LiveCodeBench (execution signals available, generated tests imperfect), MATH Level-5 hard subjects (answers normalizable or symbolically checkable), GPQA-Diamond (a low-recoverable, low-diversity boundary), and HumanEval+ (a saturated code anchor) \citep{jain2024livecodebench,hendrycks2021math,rein2024gpqa,liu2023evalplus}.

Figure~\ref{fig:framework} summarizes the fixed-pool accounting and turns it into a pre-deployment decision workflow.

Our contributions are:
\begin{enumerate}
    \item A training-free decomposition of selector/verifier gain into recoverable mass, signal coverage, conditional quality, and harm, making collaboration gains measurable at the task level rather than attributed to agent count or topology.
    \item A three-seed LiveCodeBench selector ladder comparing a first-sample baseline, LLM selectors, a generated-test verifier, a public-test verifier, and an any-of-$k$ oracle upper bound on a unified task set with hierarchical confidence intervals.
    \item Evidence that verifier fidelity is an empirical bottleneck (Figure~\ref{fig:lcb-fidelity}): a public-test verifier with high candidate-level fidelity captures much more of the oracle gap than a generated-test verifier with low fidelity, while the latter has near-zero harm.
    \item Cross-task boundary evidence: on MATH, a symbolic answer-equivalence selector improves over self-consistency while natural-language LLM selectors are negative; on GPQA-Diamond, the candidate pool has only a 3.03pp oracle gap with high answer homogeneity, and LLM selectors are net-negative on that pool.
    \item A routing-$\eta$ extension (supplementary) showing that worker routing and candidate selection obey different constraints and should not be conflated under a single notion of collaboration.
\end{enumerate}

Together, these results yield a practical diagnostic: before deploying a critic, verifier, or multi-agent selector, first estimate the oracle gap of the candidate pool, then measure whether the available signal has enough coverage, fidelity, and low harm to justify the extra computation.

\section{Related Work}

\subsection{Test-Time Selection and Execution-Based Filtering}

Self-consistency and best-of-$N$ sampling improve reasoning when aggregation or selection is reliable \citep{wang2023selfconsistency,brown2024monkeys,snell2024scaling}. For code, CodeT filters candidates with generated tests \citep{chen2022codet}, AlphaCode combines sampling with test filtering and clustering \citep{li2022alphacode}, and MBR-Exec selects by behavioral agreement \citep{shi2022mbr}. These are direct predecessors of our generated- and public-test mechanisms.

Rather than introduce another selector, we fix candidate pools and decompose realized gain into recoverable mass, signal coverage, conditional quality, and harm. This diagnostic layer directly compares natural-language, generated-test, public-test, and symbolic-equivalence mechanisms that are usually evaluated separately.

\subsection{Trained and Imperfect Verifiers}

Verifier-based selection includes trained outcome and process verifiers \citep{cobbe2021verifiers,uesato2022process,lightman2023verify}. Multi-Agent Verification introduces BoN-MAV \citep{lifshitz2025mav}, while newer general-purpose and multi-sequence verifiers improve granularity, calibration, and budget-aware ranking \citep{kwok2026llmverifier,kim2026multisequence}. These improve the verifier; our training-free setting asks when an available signal converts candidate diversity into net gain at all.

A complementary theoretical line derives best-of-$N$ and rejection-sampling behavior from verifier ROC geometry \citep{dorner2025roc}, shows that false positives impose accuracy ceilings and can bend scaling curves downward \citep{stroebl2024limits}, and establishes asymptotic advantages for verifier-based scaling \citep{setlur2025suboptimal}. Compute-matched verification is not uniformly optimal \citep{singhi2025solveverify}; \citet{venktesh2025trust} survey the broader design landscape.

Most directly, \citet{lu2025verification} systematically study when solution verification pays off across self-, within-family, and cross-family solver--verifier pairs. They find stronger cross-family verification as solver--verifier similarity decreases; our L3 instead uses a smaller cross-family selector, so the L1--L3 gap reflects capability rather than family similarity, which our capture/harm decomposition makes explicit.

We make verifier quality observable through candidate-level agreement with official labels, including accuracy, error rates, and MCC. Separating fidelity from coverage and harm captures verifiers that are accurate but rarely active, or conservative and low-harm yet miss many correct candidates, complementing imperfect-verifier theory with cross-domain empirical accounting.

\subsection{When Multi-Agent Collaboration Helps}

Recent work directly challenges the assumption that increasing agent count or interaction rounds is intrinsically beneficial. Diversity-based analysis shows that homogeneous agents saturate because their outputs are correlated, whereas heterogeneous agents provide complementary effective channels \citep{yang2026diversity}. The closest same-question work finds that entropy dynamics vary with task and coordination topology; reducing interaction-induced uncertainty therefore need not by itself establish genuine improvement \citep{zhao2026uncertainty}. Uncertainty decomposition in multi-agent debate similarly distinguishes epistemic gain from aleatoric cost \citep{qiao2026epistemic}.

System-level studies reach compatible conclusions from different angles. MAST catalogs system-design issues, inter-agent misalignment, and task-verification failures \citep{cemri2025fail}; controlled scaling studies find capability saturation and topology-dependent error propagation \citep{kim2025science}; and matched-token experiments show that single-agent reasoning can outperform multi-agent systems when total thinking compute is held fixed \citep{tran2026single}. Earlier work also finds that self-correction is unreliable without external feedback and that debate can improve some factual and reasoning tasks \citep{huang2024selfcorrect,du2023debate}.

Our scope is narrower and more mechanistic than these topology-, entropy-, or system-level analyses. We ask whether a concrete selector can convert a \emph{fixed} candidate pool into a better output. GPQA-Diamond provides a boundary case: most $k=5$ pools are answer-identical and the oracle gap is only 3.03pp, leaving little room for any selector. LiveCodeBench and MATH provide the complementary case: recoverable candidates exist, but gains depend on whether the selection signal is faithful and low-harm. Thus our oracle-gap and signal-fidelity framework is complementary to diversity and entropy: it quantifies the improvement space and the mechanism's ability to capture it.

\subsection{Generated Tests and Label-Free Fidelity Estimation}

Generated-test methods are especially close to our deployable code verifier. CoSPlay co-evolves code candidates and self-generated unit tests without ground-truth tests \citep{hu2026cosplay}. UTGen learns error-revealing inputs and expected outputs, while UTDebug spends test-time compute to validate feedback and avoid overfitting \citep{prasad2025utgen}. UTRL trains test and code generators adversarially so that generated tests better discriminate faulty programs \citep{lee2026utrl}. These approaches address the limitation surfaced by our results: self-generated tests may be low-harm yet low-fidelity because they inherit the generator's blind spots or assign incorrect expected outputs.

Our contribution is complementary. We use official labels to audit generated-test verdicts directly, identifying fidelity and activation as separate bottlenecks. This creates a bridge to label-free reliability estimation: future generated-test systems should be evaluated not only by final pass@1, but also by whether their internal confidence or reliability scores predict official-label fidelity.

\subsection{Positioning Summary}

Together, these literatures cover candidate generation, verifier training, imperfect-verifier theory, multi-agent scaling, and methods for improving generated tests. Our contribution is the missing fixed-pool measurement layer across benchmarks with different decidability profiles: before attributing gains to collaboration or topology, measure whether recoverable answers exist and whether the deployed signal has enough coverage, fidelity, quality, and low harm to recover them.

\section{Method and Experimental Setup}

\subsection{Problem Setup}

For each input $x$, a model produces a fixed candidate pool
\begin{equation}
C(x)=\{y_1,y_2,\ldots,y_k\}.
\end{equation}
The first candidate $y_1$ is the first-sample baseline. A training-free collaboration mechanism then selects one candidate or supplies a verification signal over the pool. This setup covers best-of-$N$ selection, self-consistency, LLM-based selectors, generated-test filtering, public-test filtering, and symbolic answer-equivalence selection.

We distinguish fixed-pool selection from repair. A repair mechanism may create a new answer outside the original pool, so its gain depends jointly on detection and generation. A selector can only capture improvement already present among the sampled candidates. Fixing the pool isolates the question studied here: how much available oracle space does a concrete selection mechanism convert into realized gain?

\subsection{Oracle Gap and Recoverable Mass}

For each task, the any-of-$k$ oracle succeeds if at least one candidate passes the official evaluator. Relative to a reference output, a task is \emph{recoverable} when the reference is wrong but some candidate is correct:
\begin{equation}
\mathrm{recoverable}(x)=\mathrm{reference\ wrong}(x)\wedge
\mathrm{any@}k\mathrm{\ correct}(x).
\end{equation}
The recoverable mass is $P(\mathrm{recoverable})$. The \emph{oracle gap} is the score difference between the any-of-$k$ oracle and the chosen reference. It is an upper bound on every fixed-pool selector. If the gap is small, elaborate collaboration has little absolute room to improve, regardless of selector sophistication.

\subsection{Coverage, Quality, Harm, and Fidelity}
\label{sec:decomposition}

We decompose gain using three conditional quantities:
\begin{align}
\mathrm{coverage} &= P(\mathrm{signal\ defined}),\\
\mathrm{quality} &= P(\mathrm{selected\ passes}\mid \mathrm{recoverable},\mathrm{defined}),\\
\mathrm{harm} &= P(\mathrm{selected\ fails}\mid \mathrm{reference\ passes},\mathrm{defined}).
\end{align}
\emph{Signal defined} means the mechanism produced usable evidence: for example, a parseable LLM selector output, executable public tests, generated tests that can be run, or a symbolic answer cluster. These quantities deliberately live on different slices. Quality is neither overall accuracy nor one minus harm; it asks whether the mechanism selects a correct candidate among recoverable rows where evidence is available. Harm asks whether the same mechanism turns a reference-correct row into an incorrect selected output, again conditional on evidence being defined. With these definitions, task-level gain follows Equation~\ref{eq:decomposition}.

For deployed mechanisms with a fallback policy, we additionally report \emph{effective capture} and \emph{effective harm} over all recoverable and reference-correct rows, respectively. These effective quantities include fallback and no-op behavior and therefore close the fixed-minus-harmed accounting identity. They must not be confused with the conditional quality and harm terms above. This distinction is important for conservative verifiers: a method may have low effective harm because it frequently falls back to $y_1$, even though its active signal is available on only a small slice.

We separately measure verifier \emph{fidelity} as a property of the evidence itself: candidate-level agreement between verifier verdicts and official candidate labels, summarized by accuracy, false-positive rate, false-negative rate, and MCC. Keeping fidelity separate from coverage and effective behavior prevents a high-fallback mechanism from appearing reliable merely because it seldom changes the reference.

\subsection{Selector Ladder}

We compare the mechanisms in Table~\ref{tab:tiers} on fixed $k=5$ pools. Self-consistency and majority voting are aggregation baselines rather than ladder tiers, reported where answers are naturally normalizable (MATH and GPQA). Verifier implementation depends on the benchmark: executed tests for code, symbolic or normalized-answer equivalence for MATH, and no natural executable verifier for GPQA, which therefore serves primarily as a low-recoverable boundary case.

\begin{table}[t]
\centering
\small
\caption{Selector ladder used on fixed $k=5$ pools.}
\label{tab:tiers}
\begin{tabular}{@{}lp{0.64\columnwidth}@{}}
\toprule
Tier & Description \\
\midrule
sample0 & First generated candidate. \\
L1 & Same-family LLM selects without editing. \\
L3 & Different-family or different-size selector. \\
L4-gen & Selection using generated tests. \\
L4-public & Selection using public or visible tests. \\
Oracle & Any officially correct candidate (upper bound). \\
\bottomrule
\end{tabular}
\end{table}

\subsection{Critic Actions and Routing Boundary}

The word ``critic'' can denote scientifically different actions. A verdict critic judges one candidate; a selector critic chooses among candidates; neither creates a new answer. A repair critic rewrites an answer, and test-feedback repair rewrites using an external failure signal. Only the first two are bounded by the fixed-pool oracle gap. Our main tables therefore evaluate selector and verifier mechanisms, not repair.

Worker routing is also distinct from candidate selection: it chooses a worker from input-level features before candidate evidence exists. Because routing depends on feature predictability rather than candidate-level verification, its full definition and results are reported as a supplementary boundary analysis (Appendix~A.2), not as a selector-ladder tier.

\subsection{Benchmarks, Models, and Statistical Reporting}

\begin{table}[t]
\centering
\small
\caption{Benchmarks and their diagnostic roles.}
\label{tab:benchmarks}
\begin{tabular}{@{}lrl@{}}
\toprule
Benchmark & Size & Role \\
\midrule
LiveCodeBench & 1,055 & Main code benchmark \\
MATH L5 hard & 250/seed & Symbolic answer checks \\
GPQA-Diamond & 198/seed & Low-recoverable boundary \\
HumanEval+ & 164 & Saturated code anchor \\
\bottomrule
\end{tabular}
\end{table}

The main model source is Qwen3.6-35B-A3B-BF16 for LiveCodeBench, MATH, and the GPQA pool-boundary table; Qwen3.5-9B, Qwen2.5-14B, DeepSeek-R1-Distill-Qwen-14B, and Gemma-3-4B-it appear only in follow-up, robustness, or routing analyses. To avoid provenance mismatches, every horizontally compared number carries a five-part provenance tuple $(\mathrm{benchmark},\mathrm{task\ set},\mathrm{reference},\mathrm{label\ source},\mathrm{model\ source})$; numbers with different provenance are never merged into one table, and aggregation and confidence-interval methods are declared per table. The legacy suffix \texttt{\_vllm} is a run label mapping to Qwen3.6-35B-A3B-BF16 via the run-label registry.

Code tasks use the official LiveCodeBench evaluator; MATH uses trusted regrading or explicitly marked answer-grounded checks; GPQA compares extracted option letters with dataset labels. The LiveCodeBench main table uses a three-seed all-tier common task set with task-cluster hierarchical bootstrap confidence intervals (resampling tasks, keeping seed observations within clusters). Descriptive boundary analyses (GPQA) report per-seed ranges and explicit denominators. Before running the full ladder we preregistered: (i) 8--14\% recoverable mass for the Qwen3.6-35B pool; (ii) public/execution selection beats generated-test selection; (iii) generated-test selection beats natural-language selection; (iv) cross-model natural-language selection does not stably approach execution selection; and (v) all mechanism results report coverage, quality, harm, fixed/harmed counts, confidence intervals, and common denominators. Section~\ref{sec:results} reports which expectations held.

\section{Results}
\label{sec:results}

\subsection{LiveCodeBench: Fidelity Explains the Gap between Test-Based Selectors}

Table~\ref{tab:lcb-ladder} shows the three-seed ladder on the all-tier common set (969/960/959 task-seed observations per seed; 2,888 total).

\begin{table*}[t]
\centering
\small
\caption{LiveCodeBench selector ladder on the three-seed common set.}
\label{tab:lcb-ladder}
\begin{tabular}{@{}lrrrr@{}}
\toprule
Mechanism & Pass@1 & Gain vs. sample0 & 95\% CI & Fixed / harmed \\
\midrule
First sample & 2089/2888 = 72.33\% & 0 & --- & --- \\
L3 cross-model LLM & 2146/2888 = 74.31\% & $+1.97$pp & $[0.74,3.20]$ & 175 / 118 \\
L4-gen generated tests & 2167/2888 = 75.03\% & $+2.70$pp & $[2.02,3.43]$ & 80 / 2 \\
L1 same-family LLM & 2190/2888 = 75.83\% & $+3.50$pp & $[2.26,4.73]$ & 199 / 98 \\
L4-public public tests & 2324/2888 = 80.47\% & $+8.14$pp & $[6.99,9.36]$ & 235 / 0 \\
Oracle any@5 & 2428/2888 = 84.07\% & $+11.74$pp & --- & 339 / 0 \\
\bottomrule
\end{tabular}
\end{table*}

The oracle any@5 row is not deployable; it measures recoverable mass: 339/2888 first-sample failures have a correct candidate. This bound is large enough for selection to matter, but mechanisms capture it unevenly.

\begin{figure*}[t]
\centering
\includegraphics[width=0.94\textwidth]{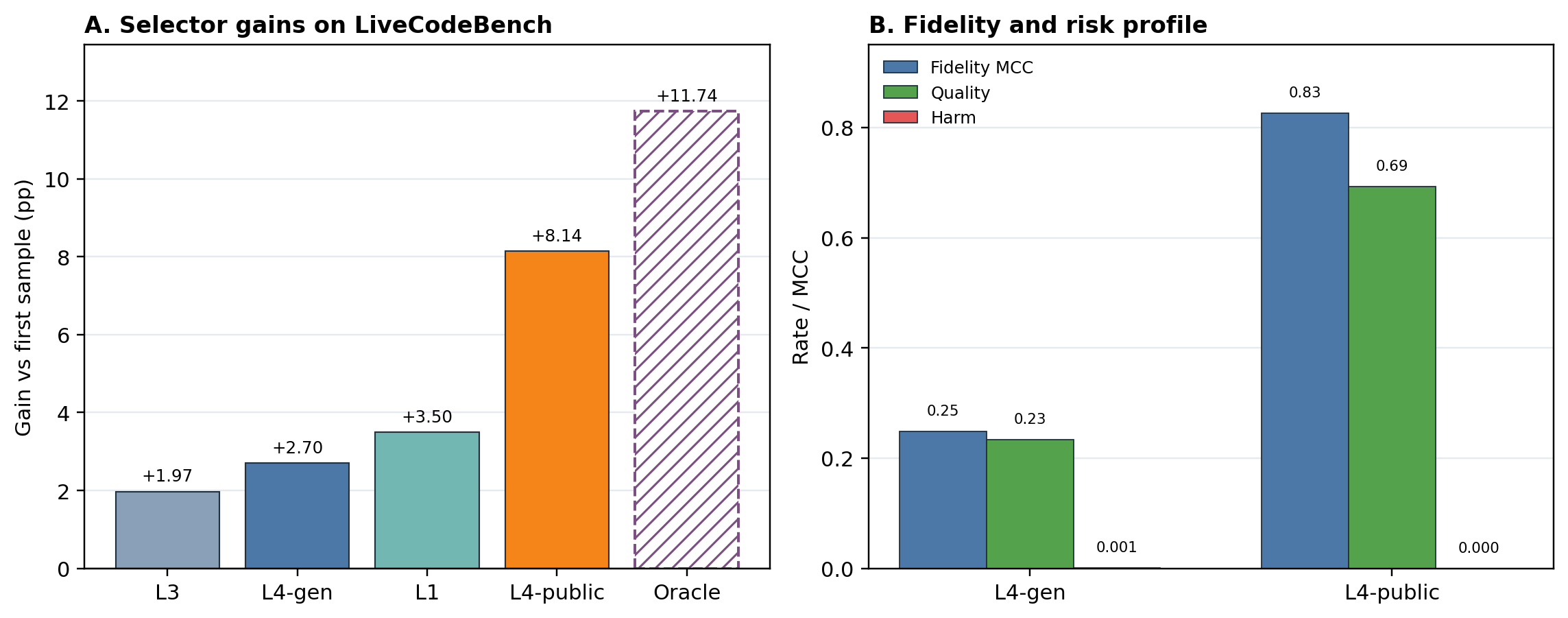}
\caption{LiveCodeBench selector performance and verifier fidelity/risk profile on the common slice. Public tests capture substantially more of the oracle gap than generated tests; generated tests remain conservative and low-harm. The hatched, dashed oracle bar is a non-deployable upper bound, visually separated from deployable mechanisms.}
\label{fig:lcb-fidelity}
\end{figure*}

The strongest contrast is between the two execution-based verifiers: $+8.14$pp (public tests) versus $+2.70$pp (generated tests). We therefore measure fidelity at candidate level. On the all-tier common candidate slice (14,440 candidates), the public-test verifier reaches 92.98\% accuracy and MCC 0.825; the generated-test verifier reaches 53.05\% and MCC 0.248 (same ordering on the full slice). Figure~\ref{fig:lcb-fidelity} visualizes the ladder and the fidelity/risk profile. The L4-public/L4-gen gap is not simply ``tests help''; it is a fidelity gap between two test signals.

The natural-language results are more nuanced. L1 is numerically higher than L4-gen by $+0.80$pp, but the direct paired contrast crosses zero (95\% CI $[-0.38,2.01]$): the preregistered prediction that generated tests beat a same-family LLM selector is not supported. The two mechanisms occupy different risk points, however: L1 captures 199/339 recoverable cases but harms 98/2089 first-sample-correct cases (4.69\%); L4-gen captures 79/339 on the clean decomposition slice with 2/2089 harm (0.10\%). The low harm of L4-gen is partly structural: generated tests exist for 2444/3165 = 77.22\% of task-seed rows, and at least one candidate passes them on only 1132/3165 = 35.77\%; a raw trace audit (2705 available rows) records 1736 fallback-to-sample0 decisions, including 618 rows with no generated tests, versus 969 active selections. The conservative fallback protects correct first samples but misses many recoverable cases.

Table~\ref{tab:accounting} instantiates the clean-slice accounting with the availability audit. Capture and harm are \emph{effective} quantities after each mechanism's fallback policy, not the conditional terms of Section~\ref{sec:decomposition}; the fixed-minus-harmed accounting identity closes exactly on the clean slice (2,887 rows; one task-seed with inconsistent timeout/harness labels is excluded from the decomposition slice only).

\begin{table*}[t]
\centering
\small
\caption{Clean-slice accounting and signal availability.}
\label{tab:accounting}
\resizebox{\textwidth}{!}{%
\begin{tabular}{@{}lrrrrr@{}}
\toprule
Mechanism & Signal availability audit & Effective capture & Effective harm & Accounting & Observed \\
\midrule
L1 & 3155/3165 = 99.68\% parseable & 199/339 = 58.70\% & 98/2089 = 4.69\% & $+3.50$pp & $+3.50$pp \\
L3 & 1671/1671 = 100\% available traces & 175/339 = 51.62\% & 118/2089 = 5.65\% & $+1.97$pp & $+1.97$pp \\
L4-gen & 77.22\% tests; 35.77\% active & 79/339 = 23.30\% & 2/2089 = 0.10\% & $+2.67$pp & $+2.67$pp \\
L4-public & 2711/3165 = 85.66\% active & 235/339 = 69.32\% & 0/2089 = 0\% & $+8.14$pp & $+8.14$pp \\
\bottomrule
\end{tabular}%
}
\end{table*}

Signal availability binds, especially for generated tests. The L4-gen gap has two separable sources: coverage/activation loss (rows falling back without active test evidence) and fidelity loss (tests disagreeing with official labels when they exist). This is why the framework keeps coverage, fidelity, effective capture, and harm separate.

Table~\ref{tab:scorecard} reports the preregistered scorecard; it is mixed, which is exactly why the decomposition is useful.

\begin{table*}[t]
\centering
\footnotesize
\caption{Preregistered LiveCodeBench scorecard.}
\label{tab:scorecard}
\setlength{\tabcolsep}{3pt}
\begin{tabular}{@{}p{0.185\textwidth}p{0.185\textwidth}p{0.185\textwidth}p{0.185\textwidth}p{0.185\textwidth}@{}}
\toprule
Recoverable mass (8--14\%) & Public $>$ generated & Generated $>$ natural language & L3 not near execution & Full reporting \\
\midrule
\shortstack[l]{Pooled 11.74\% \\ \textbf{Yes}} &
\shortstack[l]{$+8.14$ vs. $+2.70$pp \\ \textbf{Yes}} &
\shortstack[l]{Beats L3, not L1 \\ \textbf{No}; CI overlaps 0} &
\shortstack[l]{$+1.97$ vs. $+8.14$pp \\ \textbf{Yes}} &
\shortstack[l]{Tables~\ref{tab:lcb-ladder}--\ref{tab:accounting} \\ \textbf{Yes}} \\
\bottomrule
\end{tabular}
\end{table*}

\subsection{MATH: Symbolic Answer Equivalence Beats Self-Consistency}

MATH offers a different decidable signal: final answers can be normalized or checked for symbolic equivalence. Table~\ref{tab:math} reports 750 task-seed observations (250 tasks $\times$ 3 seeds) under trusted per-task regrading.

\begin{table}[t]
\centering
\small
\caption{MATH selector results under trusted regrading.}
\label{tab:math}
\begin{tabular}{@{}lrrrr@{}}
\toprule
Mechanism & Ref. & Net gain & 95\% CI & F/H \\
\midrule
L4-symbolic & SC & $+4.67$ & $[2.93,6.40]$ & 39/4 \\
L4-symbolic & sample0 & $+6.00$ & $[3.87,8.13]$ & 57/12 \\
L1 & SC & $-3.20$ & $[-5.20,-1.20]$ & 17/41 \\
L3 & SC & $-1.87$ & $[-3.73,0.00]$ & 20/34 \\
\bottomrule
\end{tabular}
\end{table}

L4-symbolic should be read carefully: it is not a problem-grounded verifier that proves candidates from the question. It clusters final answers by \texttt{math\_verify} equivalence and selects from the strongest class---518/750 = 69.07\% of rows use an equivalence key, while 232/750 = 30.93\% fall back to the first sample. The result shows that symbolic answer equivalence is a stronger aggregation baseline than surface normalized-answer SC, not that an arbitrary extra verifier beats SC.

This result does not imply that every explicit math verifier beats self-consistency. A separate L4-grounded artifact, evaluated under its own label source, selects 419/750 correct answers while SC selects 420/750. It fixes no SC failures and harms one SC success, for $-0.13$pp relative to SC under that artifact. We therefore treat L4-grounded as evidence that a weak grounding signal may fail to exceed SC, while the trusted L4-symbolic result is the formal positive result. Together, the MATH results reinforce the framework's main distinction: an extra natural-language judge is not enough, but a high-fidelity answer-equivalence signal can convert recoverable mass into gain.

\subsection{GPQA-Diamond: The Oracle Gap Is Too Small to Pay for Harm}

GPQA-Diamond answers normalize to option letters, but correctness is not execution-checkable. Table~\ref{tab:gpqa} reports the three-seed boundary on the Qwen3.6-35B pool.

\begin{table}[t]
\centering
\small
\caption{GPQA-Diamond boundary on the 35B pool.}
\label{tab:gpqa}
\begin{tabular}{@{}lrrr@{}}
\toprule
Mechanism & Accuracy & Gain & Fixed/harmed \\
\midrule
First sample & 283/594 = 47.64\% & 0 & --- \\
SC majority & 273/594 = 45.96\% & $-1.68$pp & --- \\
L1 & 273/594 = 45.96\% & $-1.68$pp & 5/15 \\
L3 & 269/594 = 45.29\% & $-2.36$pp & 6/20 \\
Oracle any@5 & 301/594 = 50.67\% & $+3.03$pp & --- \\
\bottomrule
\end{tabular}
\parbox{0.96\columnwidth}{\footnotesize\emph{Note:} Across the three seeds, row-wise accuracy ranges are $[45.96,48.99]$, $[43.94,47.98]$, $[43.94,47.98]$, $[43.94,46.97]$, and $[49.49,52.53]$, respectively.}
\end{table}

The oracle gap is only 18/594 = 3.03pp: a perfect selector could add about three points. The raw pools explain why: 520/594 pools (87.54\%) contain five identical answer letters (mean unique letters 1.138). SC is 10/594 below the first sample under the pipeline's tie-break rule. LLM selectors are net-negative, and predictably so: L1 captures 5/18 recoverable cases but harms 15/283 first-sample-correct cases (5.30\% harm; L3: 6/18, 7.07\%). The cost side is roughly first-sample-correct mass times harm rate---for L1, 15/594 = 2.53pp, already exceeding what five recovered cases can compensate. L1 and SC tie at 273/594 by cancellation, not degeneration: they disagree on 20 observations, each uniquely correct on 10. A weaker Qwen3.5-9B pool shrinks the boundary further (recoverable 0.67\%, 94.44\% identical pools): the oracle gap is a joint property of task, model, and sampling configuration, not a benchmark constant.

\subsection{Summary}

The same diagnostic pattern appears across benchmarks---and signal availability binds across them as well (35.77\% active generated-test rows on LiveCodeBench; 69.07\% equivalence-key rows on MATH). On LiveCodeBench, substantial recoverable mass exists and high-fidelity public tests capture much of it; generated tests are positive but limited by low fidelity and low activation. On MATH, symbolic answer equivalence beats SC while LLM selectors are negative for L1 and marginal for L3. On GPQA-Diamond, the pool has too little recoverable mass for LLM selector harm to be worthwhile. Training-free collaboration is not intrinsically useful because it adds another agent; it is useful when the pool contains recoverable answers and the signal has enough coverage, fidelity, quality, and low harm.

\section{Discussion and Limitations}

\subsection{Collaboration Gain Is Not a Topology Property}

The LiveCodeBench contrast among L4-public, L4-gen, L1, and L3 shows that gains do not arise from adding a judging layer by itself. Public tests are strongest, but the deployable generated-test verifier is not statistically distinguishable from L1 in this setting. The relevant axis is not simply test-based versus natural-language selection; it is the joint effect of signal fidelity, coverage, conditional quality, fallback behavior, and harm. The large L4-public--L4-gen gap identifies generated-test fidelity and activation as the current bottlenecks.

\subsection{Best-of-\texorpdfstring{$N$}{N} Supplies Candidate Space}

Best-of-$N$ and self-consistency are not outside collaboration. They create the candidate pool; selectors and verifiers determine whether that pool is converted into a better final answer. Without an oracle gap, no fixed-pool selector can help. With an oracle gap but weak verification, the available improvement remains uncaptured. This two-stage view separates candidate generation from selection and makes clear why agent count or sampling budget alone is not an explanation of realized gain.

\subsection{Scope and Limitations}

First, LiveCodeBench selected-output evaluation and stored candidate oracle labels contain a small timeout/harness inconsistency. We identify \texttt{seed43/lcb\_abc323\_d} as the responsible task-seed and exclude it only from the clean decomposition slice, while retaining official selected-output evaluation in the main pass@1 table.

Second, candidate-level fidelity is fully measured for L4-public and L4-gen, not for the L1/L3 natural-language selectors. The L3 availability audit also covers only the available trace subset rather than the full selected-output slice. Adding LLM verdict matrices would support a broader fidelity comparison; we leave this extension for future work rather than treating final pass@1 as a substitute for signal fidelity.

Third, L4-public uses task-visible tests and should be read as a public-test verifier or partial-oracle diagnostic upper bound, not as a fully deployable mechanism. The current overlap audit is a 50-task sample rather than a complete 1,055-task non-overlap proof.

Fourth, MATH contains two mechanism and label-source regimes. The trusted L4-symbolic per-task selector is the formal positive result. L4-grounded uses a separate grounded-summary label source and is interpreted separately rather than merged into the same horizontal comparison.

Fifth, GPQA-Diamond establishes a low-recoverable boundary for the studied Qwen3.6-35B pool, not for every natural-language knowledge task. The oracle gap is a joint property of the task, model, prompt, temperature, and sampling configuration; more diverse sampling could change the available recoverable mass.

Sixth, this paper studies selector and verifier mechanisms, not repair. Repair critics can create answers outside the fixed pool, but they introduce a distinct detection--generation trade-off and a different harm profile.

Seventh, we do not claim compute optimality. All main selector results use $k=5$ candidate pools, and the LiveCodeBench token-cost proxy was not recomputed on the final all-tier common set. Earlier probes suggest comparable gain per selector token for L4-gen and L1, with lower efficiency for L3, but those are appendix-level diagnostics rather than a compute-matched conclusion. A deployment decision should compare verifier calls with larger-$k$ sampling, normalized self-consistency where applicable, and execution cost \citep{snell2024scaling,singhi2025solveverify}.

Finally, the diagnostic requires labels on a development set. Estimating oracle gap, fidelity, quality, and harm is not label-free in the form used here; it is intended as a small labeled audit before investing in a collaboration mechanism. GPQA illustrates the potential sample efficiency of that audit: 594 task-seed observations are sufficient to identify a 3.03pp oracle gap and predict net-negative LLM selection. Label-free reliability estimation for generated tests is a promising substitute when development labels are unavailable.

\section{Conclusion}

Training-free collaboration is not valuable merely because it adds an agent or critic. In fixed-pool selection, recoverable mass bounds gain; signal coverage, fidelity, quality, and harm determine how much is realized. High-fidelity public tests capture substantial oracle space on LiveCodeBench, symbolic equivalence beats self-consistency on MATH, and LLM selectors become net-negative on homogeneous GPQA-Diamond pools with a 3.03pp oracle gap. The deployment rule is therefore: estimate oracle gap first, then audit signal availability, fidelity, and harm before spending additional inference compute.

\subsubsection*{Reproducibility Statement}
The paper reports task-set denominators, seed counts, model provenance, label sources, selection rules, fixed/harmed counts, and confidence-interval procedures. The supplementary material documents the routing extension and additional audit details. Code and artifacts, including the run-label registry, selected-output records, candidate verdicts, and analysis scripts, are available at \url{https://github.com/AmGarfield/OracleGap}.

\bibliography{references}
\bibliographystyle{oraclegap_arxiv}

\appendix

\section{Supplementary Results}

\subsection{HumanEval+ Saturation Anchor}

HumanEval+ serves only as a saturated code anchor. In the Qwen3.6-35B V100 follow-up, first-sample pass@1 is 138/164, 139/164, and 138/164 across seeds 43/44/45, while oracle any@5 is 144/164, 144/164, and 142/164. The corresponding recoverable counts are only 6/164, 5/164, and 4/164, so HumanEval+ is not used as the main code selector benchmark.

\subsection{Routing-\texorpdfstring{$\eta$}{eta} Extension}

Worker routing chooses which model should answer an input rather than selecting among candidates from a fixed pool. We define
\begin{equation}
\eta = \frac{\mathrm{score}(\mathrm{learned\ router})-\mathrm{score}(\mathrm{default\ worker})}
{\mathrm{score}(\mathrm{oracle\ router})-\mathrm{score}(\mathrm{default\ worker})}.
\end{equation}
An $\eta$ near one means that the learned router captures most of the oracle routing space; an $\eta$ near zero means it does not exploit the available worker differences. Table~\ref{tab:routing} summarizes the strongest-feature runs. These results are descriptive and are not part of the selector-ladder evidence because the default-worker identities and confidence-interval pipeline are not integrated into the paper-level provenance system.

\begin{table}[h]
\centering
\small
\caption{Routing-$\eta$ extension.}
\label{tab:routing}
\begin{tabular}{@{}llrrrr@{}}
\toprule
Benchmark & Router & Default & Router & Oracle & $\eta$ (95\% CI) \\
\midrule
LCB & Logistic & 64.93 & 64.93 & 69.48 & 0.000 $[0,0]$ \\
LCB & Random forest & 64.93 & 66.26 & 69.48 & 0.292 $[.122,.459]$ \\
MATH & Logistic & 46.80 & 58.80 & 60.00 & 0.909 $[.783,1]$ \\
MATH & Random forest & 46.80 & 58.00 & 60.00 & 0.848 $[.700,.968]$ \\
GPQA & Logistic & 43.94 & 56.57 & 66.16 & 0.568 $[.368,.750]$ \\
GPQA & Random forest & 43.94 & 57.07 & 66.16 & 0.591 $[.386,.769]$ \\
\bottomrule
\end{tabular}
\end{table}

The LCB logistic router degenerates to selecting the default worker, yielding $\eta=0$. The remaining results show that input-level routing can recover some oracle-routing space, but this depends on worker differences and feature predictability rather than candidate-level verification. Candidate selection and worker routing are both test-time collaboration mechanisms, but they obey different constraints.

\subsection{Additional Audit Details}

One task-seed observation, \texttt{seed43/lcb\_abc323\_d}, is excluded from the clean decomposition slice because selected-output evaluation marks it as passing while all stored candidate oracle labels fail with timeout or harness codes. The main pass@1 table retains the official selected-output evaluation. The raw generated-test trace audit contains 2,705 available rows, versus 2,887 rows in the clean decomposition slice, because the seed-45 trace artifact is incomplete; final pass@1 uses the official selected-output details.

The public-test verifier should be interpreted as a partial-oracle diagnostic upper bound, not a fully deployable mechanism. The current overlap audit covers a 50-task sample rather than proving non-overlap for all 1,055 LiveCodeBench tasks. The MATH grounded artifact uses a separate label source and is therefore not merged with the trusted symbolic-equivalence result.

\subsection{LLM Usage Disclosure}

An LLM-based assistant was used for language editing and preparation of the \LaTeX{} manuscript. The author reviewed the resulting text, equations, tables, citations, and numerical claims and retains full responsibility for the submission.

\end{document}